\documentclass[conference]{IEEEtran}
\IEEEoverridecommandlockouts
\usepackage{cite}
\usepackage{amsmath,amssymb,amsfonts}
\usepackage{algorithmic}
\usepackage{graphicx}
\usepackage{textcomp}
\usepackage{xcolor}
\usepackage{multirow}
\def\BibTeX{{\rm B\kern-.05em{\sc i\kern-.025em b}\kern-.08em
    T\kern-.1667em\lower.7ex\hbox{E}\kern-.125emX}}
\makeatletter
\newcommand{\linebreakand}{%
  \end{@IEEEauthorhalign}
  \hfill\mbox{}\par
  \mbox{}\hfill\begin{@IEEEauthorhalign}
}
\makeatother

\begin{document}

\title{Semantic Is Enough: Only Semantic Information For NeRF Reconstruction\\
\thanks{*Corresponding Author}
}

\makeatletter
\def\footnoterule{\kern-3\p@
  \hrule \@width 3.5in \kern 2.6\p@} 
\makeatother

\author{\IEEEauthorblockN{RUIBO WANG}
\IEEEauthorblockA{\textit{Smart Cloud Platform} \\
\textit{Z-one Technology Co., LTD.}\\
Shanghai, China \\
wangruibo01@saicmotor.com}
\and
\IEEEauthorblockN{SONG ZHANG*}
\IEEEauthorblockA{\textit{Auto-driving Computing Platform} \\
\textit{Z-one Technology Co., LTD.}\\
Shanghai, China \\
zhangsong05@saicmotor.com}
\and
\IEEEauthorblockN{PING HUANG}
\IEEEauthorblockA{\textit{Smart Cloud Platform} \\
\textit{Z-one Technology Co., LTD.}\\
Shanghai, China \\
hunagping01@saicmotor.com}
\linebreakand 
\IEEEauthorblockN{DONGHAI ZHANG}
\IEEEauthorblockA{\textit{Smart Cloud Platform} \\
\textit{Z-one Technology Co., LTD.}\\
Shanghai, China \\
zhangdonghai@saicmotor.com }
\and
\IEEEauthorblockN{WEI YAN}
\IEEEauthorblockA{\textit{Smart Cloud Platform} \\
\textit{Z-one Technology Co., LTD.}\\
Shanghai, China \\
yanwei04@saicmotor.com}
}

\maketitle
\begin{abstract}
Recent research that combines implicit 3D representation with semantic information, like Semantic-NeRF, has proven that NeRF model could perform excellently in rendering 3D structures with semantic labels. This research aims to extend the Semantic Neural Radiance Fields (Semantic-NeRF) model by focusing solely on semantic output and removing the RGB output component. We reformulate the model and its training procedure to leverage only the cross-entropy loss between the model's semantic output and the ground truth semantic images, removing the colour data traditionally used in the original Semantic-NeRF approach. We then conduct a series of identical experiments using the original and the modified Semantic-NeRF model. Our primary objective is to obverse the impact of this modification on the model's performance by Semantic-NeRF, focusing on tasks such as scene understanding, object detection, and segmentation. The results offer valuable insights into the new way of rendering the scenes and provide an avenue for further research and development in semantic-focused 3D scene understanding.
\end{abstract}

\begin{IEEEkeywords}
Semantic, NeRF, Denoise, Super-resolution, Label Propagation, Sparse Label
\end{IEEEkeywords}

\section{Introduction}
Recent advances in 3D understanding and computer vision have opened up new avenues for traditional ways. One vital technique, known as Neural Radiance Fields (NeRF) \cite{mildenhall2021nerf}, a compelling methodology for synthesising novel views of 3D scenes, has widespread application in computer vision and 3D scene understanding. By leveraging a fully-connected neural network, NeRF models the volumetric scene function and has achieved remarkable results in photo-realistic implicit rendering. The concept of Semantic-NeRF  \cite{zhi2021place} further enriches the NeRF model by adding semantic information to the rendering process, allowing for better scene understanding. It expands on the conventional approach by representing scenes through RGB data and incorporating semantic labels so that it extends beyond the boundaries of RGB data into the realm of semantic knowledge of the scene.
\par As in the Semantic-NeRF, it provides both geometric and semantic, which coordinate to train the network and show that semantics is highly related to geometry information. The semantic part also helps to learn the implicit 3D structure with the well-defined semantic label, so only semantic information could guide the model to produce a 3D semantic representation. In this paper, we show an approach that simplifies the architecture of Semantic-NeRF by eliminating RGB output from the equation and only concentrating on the semantic information. We achieve this by altering the training procedure of the neural network to only employ the semantic loss to update the network, thereby reducing the colour part in the network to form an integral part of the original model.
The crux of our research is to examine this modification's implications on the model's overall performance and whether the given info could provide a similar result. This RGB-less version of Semantic-NeRF departs from the standard practice. By removing colour information and focusing solely on the semantic data, the model will provide results in tasks only concerned with the 3D semantic interpretation of the scene.
\par We conduct comprehensive experiments to validate our approach and compare the original Semantic-NeRF with our modified version. The experiments are designed to provide a head-to-head comparison in identical experimental conditions in the Replica dataset \cite{straub2019replica}, such as noise, semantic accuracy, super-pixel, etc. The comparison critically assesses how removing RGB output and the exclusive use of semantic information affects the model's performance. Furthermore, the experimental comparison helps in understanding the merits and demerits of this proposed modification. As the experiment result, our approach shows a similar ability to render the semantic part with Semantic-NeRF.
\par To sum up, this research endeavours to modify the Semantic-NeRF by removing the RGB output, focusing only on the semantic production, and altering the training methodology accordingly. By comparing this modified model with the original Semantic-NeRF under the same experimental conditions. Through those experiments, we aim to elucidate the potential impact of this modification on the model's performance to contribute valuable insight into semantic 3D representation.

\section{Related Works}
MLPs-based neural networks representing 3D scenes from images have become a prominent research area in computer vision and machine learning due to their success in creating novel view synthesises and high-quality graphics. With the base of 3D geometry, some semantic methods combine to illustrate structures and semantics.
\subsection{Neural Radiance Fields method}

Neural Radiance Fields (NeRF) \cite{mildenhall2021nerf} has presented an MLPs method using images to reconstruct the 3D scene. However NeRF has problems in many aspects, such as training speed, bounded scene, multi-scaled consistency, etc. So, there are many methods to improve NeRF for better performance. NeRF lacks keeping the multi-scale image quality; Mip-nerf \cite{barron2021mip} indicates a way of using the cone-like ray field to render the image, which promises that whatever the input scale change, it will always perform reliable output. 
\par Training NeRF always consumes a long time, so those methods give a solution to speed up the training. Depth-supervised NeRF \cite{deng2022depth} using the depth map to accelerate training. The \cite{yu2021plenoctrees} and \cite{fridovich2022plenoxels} use the explicit voxel structure with spherical harmonics to reduce the training steps. Using the point cloud, \cite{xu2022point} makes the point cloud the input to simplify the MLP structure to reconstruct quickly. 
\par In the unbounded scene, some improvement methods come to the fore. \cite{zhang2020nerf++} makes the scene into foreground and background, separating each using a unit sphere and then combining them to render the unbounded scene. \cite{barron2022mip} using the scene and ray parameterisation to overcome the problems in unbounded scenes.
Other methods like \cite{lin2021barf} and \cite{meng2021gnerf} solve the problems when the camera pose is not accurate when those poses are used as the input.

\subsection{3D Semantic Representation}
Some traditional methods use the 3D reconstruction structure joint with semantic info to form the 3D semantic representation. \cite{hermans2014dense} using the point cloud to attach semantic labels, based on the SLAM method, \cite{mccormac2017semanticfusion} and \cite{runz2018maskfusion} fuse the semantic tags with SLAM reconstruction, \cite{narita2019panopticfusion} using voxels, combined with semantic map, \cite{mccormac2018fusion++} draws the semantic information on signed distance field reconstructions.
\par In the neural rendering with the semantic method, \cite{zhi2021place} uses image and semantic loss to reconstruct the colour and semantic views. \cite{fu2022panoptic} takes sparse photos, coarse 3D bounding primitives and semantic images as the input, which gives the panoptic semantic views. \cite{kundu2022panoptic} training the whole scene separately into foreground and background with semantic labels. \cite{fan2022nerf} altering the supervised learning into self-supervised learning to generate 3D semantic information. \cite{vora2021nesf} supervised by 2D semantic pose with U-net to form a 3D semantic field. \cite{zhi2021ilabel} creating a GUI, which provides human-computer interaction, allows users to focus on editing efficiently. 
\par In the single-image input 3D geometry method, \cite{kundu20183d} predicts the 3D mesh with semantics in the image. \cite{dahnert2021panoptic} shows the ability in 3D panoptic scenes with semantic labels.

\section{Method}
\subsection{Preliminaries}
NeRF \cite{mildenhall2021nerf} is a method that implicitly employs Multilayer Perceptrons (MLPs) to represent 3D structures. By utilising the intrinsic and extrinsic parameters of the camera, we can obtain the world coordinates \textit{X} = (\textit{x}, \textit{y}, \textit{z}) and view directions \textit{d} = ($\theta$, $\phi$) for the corresponding images, which are used as inputs for NeRF. These inputs are encoded via positional encoding to ensure the effects of high-frequency colour and geometry. The formula $\gamma$ is as follows:
\begin{equation}
\gamma(\textit{X}) = (sin(\textit{X}),cos(\textit{X}),...,sin(2^{L-1}\pi \textit{X}),cos(2^{L-1}\pi \textit{X}))\label{eq}
\end{equation}

\begin{figure}[t]
\centerline{\includegraphics[width=0.5\textwidth]{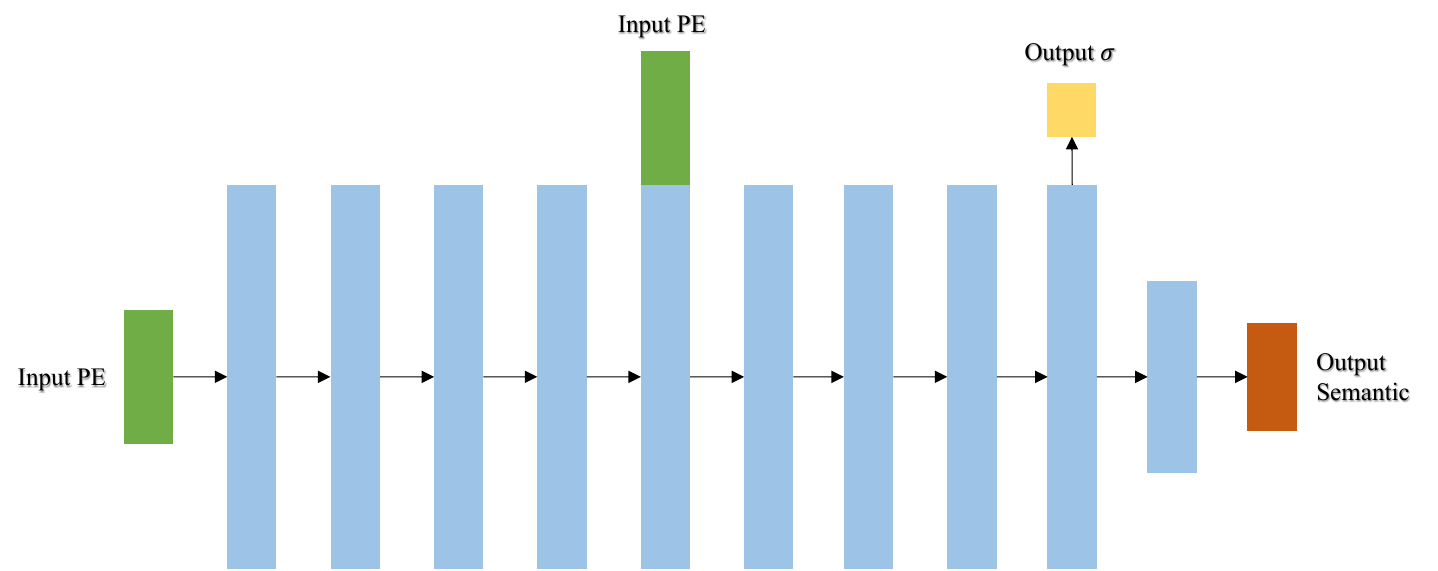}}
\caption{The network architecture of our method. This network uses the Positional encoding of the world coordinates \textit{X} = (\textit{x},\textit{y},\textit{z}) as the input to our network, and in the $5^{th}$ blue layer, it will concatenate the input into the network. Each blue layer uses 256 channels except the last one, which uses 128 channels, and the output layer uses the channel as same as the number of semantic labels. Then, our network outputs the volume density $\sigma$ in the yellow block and the semantic logit in the brown block.}
\label{fig}
\end{figure}

After computing it with sine or cosine functions, this formula concatenates the 3-dimensional positional input \textit{X}. Each \textit{X} coefficient is the 0 to L-1 power of 2 times $\pi$ during the calculation. L is the hyper-parameter that can be set.
\par The output of NeRF is colour \textit{c} = (\textit{r}, \textit{g}, \textit{b}) and volume density $\sigma$, which are subsequently used for hierarchical volume rendering. When calculating the hierarchical volume rendering, NeRF uses quadrature uniform sample rays emitted from the camera centre to the pixel points of the image. Also, it provides the bounds for near and far distances of the ray: $[t_{near},t_{far}]$, with the equation:
\begin{equation}
\textit{ray} = \textit{o} + \textit{t}\textit{d}\label{eq}
\end{equation}
Where \textit{o} is the camera centre coordinate, \textit{t} is the distance of the ray, and \textit{d} is the view direction. The corresponding pixel colour is determined through the hierarchical volume rendering formula.
\begin{equation}
\begin{aligned}
\textit{C(ray)} &= \sum_{n=1}^{N}T_{n}(1-exp(-\sigma_{n}\delta_{n}))c_{n}
\\&T_{n} = exp(-\sum_{m=1}^{n-1}\sigma_{m}\delta_{m})
\label{eq}
\end{aligned}
\end{equation}
Where $\sigma_{n}$ is the $n^{th}$ volume density, $\delta_{n}$ is the ray distance difference in $t_{n+1} - t_{n}$, $c_{n}$ is the nth ray color.
\par This network employs two parts for training, the "coarse" and the "fine" network. The "coarse" network uses the results of quadrature uniform sampling for hierarchical volume rendering, while the "fine" network conducts inverse sampling based on quadrature uniform sampling for hierarchical volume rendering.
\par Semantic-NeRF \cite{zhi2021place} adds a semantic layer to the NeRF MLPs, generating semantic label outputs. This method only uses the world coordinates \textit{X} = (\textit{x}, \textit{y}, \textit{z}) to obtain the output of the semantic layer and performs hierarchical volume rendering on the semantic layer output to get the semantic label value of the corresponding pixel in the image. When processing the semantic-layer results of Semantic-NeRF, the hierarchical volume rendering formula is modified. In the semantic layer, the pixel value corresponding to each ray in the image is a semantic label, so the volume rendering formula changes \textit{C(ray)} to \textit{S(ray)}, thus performing hierarchical volume rendering corresponding to the semantic output.

\begin{figure}[t]
\centerline{\includegraphics[width=0.5\textwidth]{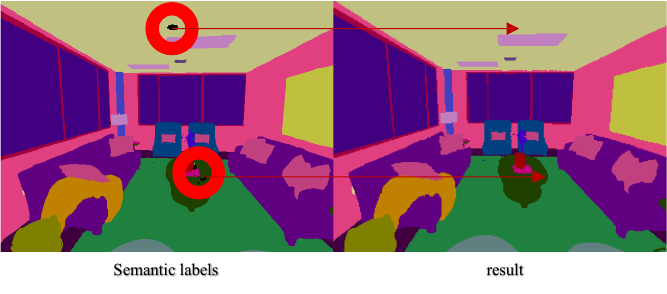}}
\caption{In our training step, we set our loss function to ignore those unnecessary labels like the black part in the red circles in the training data sets so that in both the semantic nerf method and our method, it will fill it with the nearby semantic labels, and the red arrows show the difference by ignoring those black parts.}
\label{fig}
\end{figure}

\subsection{Semantic Is Enough}
Our method proposes completing scene reconstruction and semantic output without using RGB images. Fig. 1 shows the network design of our method by using multiple perception layers with only semantic output.
\par Based on Semantic-NeRF \cite{zhi2021place}, our method uses only the world coordinates \textit{X} = (\textit{x}, \textit{y}, \textit{z}) to obtain semantic output \textit{S}:
\begin{equation}
S=MLPs(X)\label{eq}
\end{equation}
With the hierarchical volume rendering formula:
\begin{equation}
\begin{aligned}
\textit{S(ray)} &= \sum_{n=1}^{N}T_{n}(1-exp(-\sigma_{n}\delta_{n}))s_{n}
\\&T_{n} = exp(-\sum_{m=1}^{n-1}\sigma_{m}\delta_{m})
\label{eq}
\end{aligned}
\end{equation}
Where $\sigma_{n}$ and $\delta_{n}$ are the same as the Semantic-NeRF, which are the $n^{th}$ volume density and the ray distance difference, and $s_{n}$ is the nth ray semantic label.
\par Using the coarse and fine networks, we obtain rays from quadrature uniform sampling and inverse sampling based on quadrature uniform sampling. These rays project the corresponding values of each pixel on each semantic image. We adopt the hierarchical volume rendering formula of Semantic-NeRF for the semantic layer. The output results are the semantic results obtained through hierarchical volume rendering, representing the probability of the corresponding semantic label at each pixel point.
\par During network training, we calculate the cross-entropy loss between the semantic results obtained and the ground truth semantic image for network training. We only need this one loss to complete the training of the entire network. The semantic loss is defined as follows:
\begin{equation}
\begin{aligned}
Loss = -\sum_{ray\in R}[&\sum_{l=1}^{L}p^{l}(ray)log(p_{c}^{'l}(ray)) \quad + \\&\sum_{l=1}^{L}p^{l}(ray)log(p_{f}^{'l}(ray))]\label{eq}
\end{aligned}
\end{equation}
Where \textit{L} represents all semantic labels in the training set, \textit{R} represents all rays in the training set, and $p^{l}(ray)$ represents the probability of this ray corresponding to label l in the ground truth. $p_{c}^{'l}(ray)$ and $p^{l}(ray)$ represent the probability predicted by the coarse and fine networks that this ray corresponds to label l.

\begin{figure}[t]
\centerline{\includegraphics[width=0.5\textwidth]{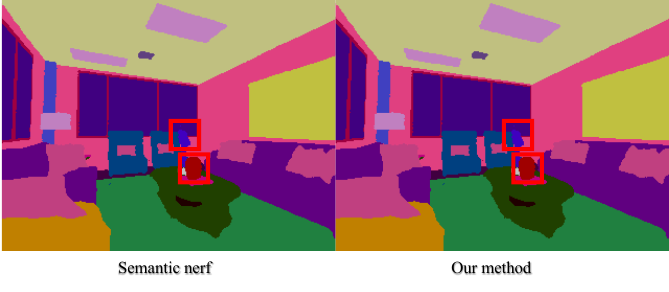}}
\caption{Qualitative comparison of semantic view synthesis. Our semantic-only method results compared to the Semantic NeRF in rendering a novel view of semantic information, it shows almost no difference between those results, and we can focus on those red rectangles, which show some specific items, our method performs similar rendering quality as Semantic NeRF.}
\label{fig}
\end{figure}

\section{Experiment}
Our entire experiment uses hyper-parameters similar to those in \cite{mildenhall2021nerf} and \cite{zhi2021place}. In positional encoding, we use encoding lengths of 10 and 4. Regarding the bounds settings for rays, we continue to follow the setting method of Semantic-NeRF from 0.1m to 10m.
For the training set in our experiment, we resize all images in the size of 320x240, which can reduce the computational cost, and we implement our method and data training using the PyTorch \cite{paszke2019pytorch} framework. The entire experiment is conducted on an Nvidia A100 80G, with a ray batch size of 1024, and we finally use Adam \cite{kingma2014adam} as the 
optimizer, setting the learning rate at 5e-4. The entire experiment will undergo 200,000 iterations.

\begin{figure}[t]
\centerline{\includegraphics[width=0.5\textwidth]{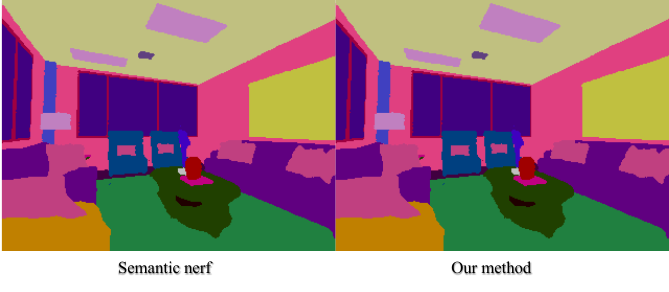}}
\caption{Qualitative comparison of the sparse label. When we use only 10\% of semantic images as the training datasets, our method also shows great power in using a small number of datasets to render the semantic maps as the Semantic NeRF.}
\label{fig}
\end{figure}
\subsection{Dataset}
Replica \cite{straub2019replica} dataset is a high-fidelity, diverse and realistic dataset with semantic labels using an AI Habitat simulator \cite{savva2019habitat} rendering images designed to advance research in embodied artificial intelligence. Whole datasets provide 18 scenes and 88 semantic labels, mapped to colour with the NYUv2-13 definition \cite{zhi2021place}, \cite{eigen2015predicting}, \cite{silberman2012indoor}.
\par Our experiment uses the scene in room 0 with sequence one, which contains 900 images in size 640x480. The pose we used directly from the pin-hole camera and the total number of semantic labels used in this scene is about 29 classes. As the semantic label appears, some part is useless in our loss computation, so we only use 28 classes for loss computation. Still, our method will automatically fill those missing parts, like in Fig. 2; this will cause an effect in calculating the MIoU and accuracy results in experiments. In splitting training and testing datasets, we process that in this way. We will concatenate them in every 5th image frame into our training dataset. And in every 5th image, which starts from the second frame, we allocate them into our testing dataset.

\begin{figure}[t]
\centerline{\includegraphics[width=0.5\textwidth]{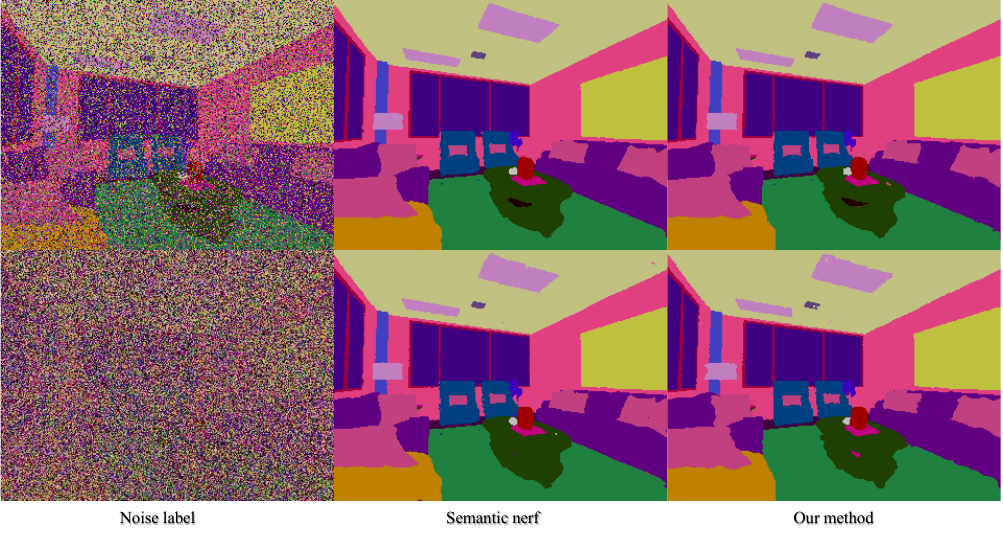}}
\caption{Qualitative comparison of the Pixel-wise noise. From different levels of pixel corruption, we can see that with the 50\% noise labels in the first row, our methods can denoise those labels to their original forms as the Semantic NeRF. Even in the 90\% noise labels in the second row, our method also performs a similar output as the Semantic NeRF.}
\label{fig}
\end{figure}

\subsection{Semantic View Synthesis}
\begin{table}[hb]
\caption{QUANTITIVE RESULT IN SEMANTIC VIEW SYNTHESIS}
\begin{center}
\begin{tabular}{|c|ccc|}
\hline
\multirow{2}{*}{\textbf{Model}} & \multicolumn{3}{c|}{\textbf{Metric}}                           \\ \cline{2-4} 
                  & \multicolumn{1}{c|}{\textbf{\textit{MIoU}}} & \multicolumn{1}{c|}{\textbf{\textit{Total ACC}}} &\textbf{\textit{AVG ACC}}   \\ \hline
                  Semantic NeRF& \multicolumn{1}{c|}{0.972} & \multicolumn{1}{c|}{0.995} &0.984  \\ \hline
                  Our Method& \multicolumn{1}{c|}{0.973} & \multicolumn{1}{c|}{0.996} & 0.986 \\ \hline
\end{tabular}
\label{tab1}
\end{center}
\end{table}
We use geometric structures to assist with semantic reconstruction. Here, we compare the differences between the semantic images produced by our method and Semantic-NeRF \cite{zhi2021place}. After both methods were learned from scratch, as seen in Fig. 3, there is no significant difference between the pure semantic method and the Semantic-NeRF method, and neither is there a substantial difference from the ground truth. The quantitative results can be seen in Table I.


\subsection{Sparse Label}
In real scenarios, obtaining the entire dataset's semantic labels is often time-consuming and costly, requiring substantial human and financial resources. Semantic-NeRF \cite{zhi2021place} suggests that extracting information from some keyframes to train the 3D semantic structure is sufficient. We train our method with the same experimental settings. Semantic-NeRF's experiment shows that using more than 10\% of the semantic keyframes does not significantly affect the final result, but the effect worsens with less than 10\%. Therefore, our experiment uses 10\% of the semantic keyframes as the training set, the same as Semantic-NeRF, to verify whether this boundary effect applies to our method. From the results of Fig. 4 and Table II, this rule still applies to our pure semantic method, and our method can still provide similar impacts to Semantic-NeRF.

\begin{figure*}[ht]
\centerline{\includegraphics[width=1\textwidth]{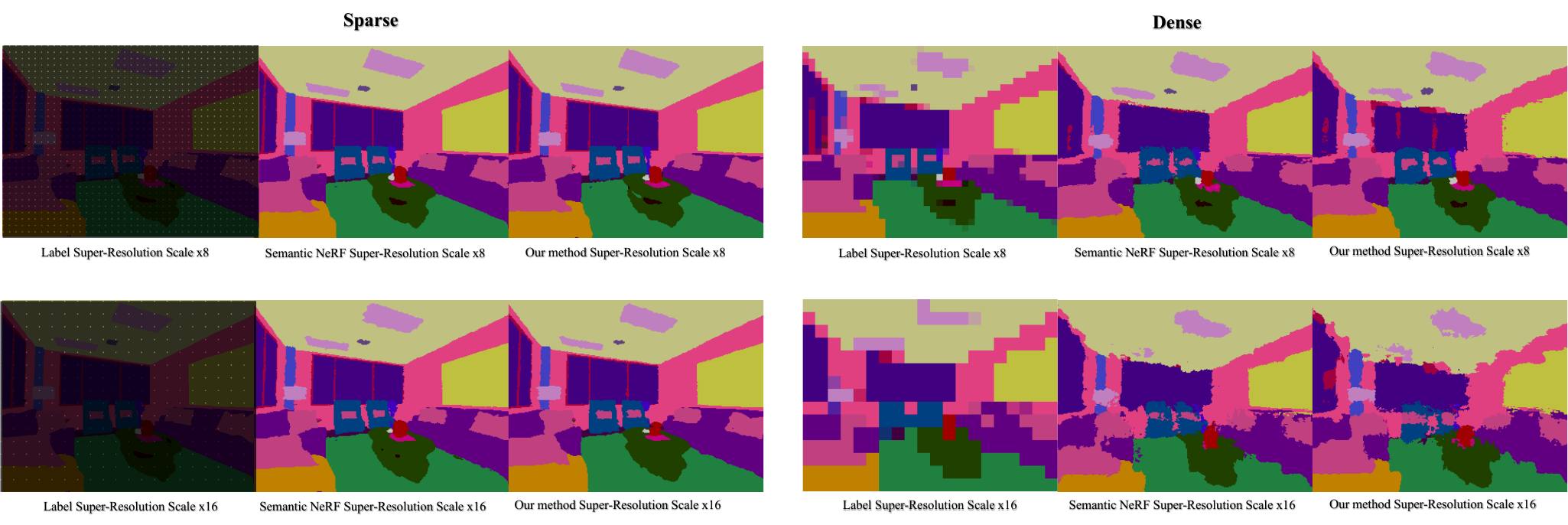}}
\caption{Qualitative comparison of super-resolution. On the left side are sparse label results. After training the input label, our method obtains a similar quality in the image with Semantic-NeRF in two scales of super-resolution. The right side is dense label results, in 8 and 16 times down-scales. Our method can perform a similar output with Semantic-NeRF, and both methods can recover the low-resolution image in a similar quality.}
\label{fig}
\end{figure*}
\subsection{Multi-view Semantic Fusion}
Multi-view consistency is an essential part of 3D reconstruction methods. The NeRF \cite{mildenhall2021nerf} method emphasises its role in multi-view consistency, and Semantic-NeRF also inherits the same feature. The multi-view consistency is also demonstrated in some traditional semantic mapping systems, such as \cite{sunderhauf2017meaningful}. Semantic-NeRF \cite{zhi2021place} trains to achieve experiments take the same settings as Semantic-NeRF, masking the same parts, including pixel-wise noise, super-resolution, and label propagation.
\begin{table}[!hb]
\caption{QUANTITIVE RESULT IN SPARSE LABEL}
\begin{center}
\begin{tabular}{|c|ccc|}
\hline
\multirow{2}{*}{\textbf{Model}} & \multicolumn{3}{c|}{\textbf{Metric}}                           \\ \cline{2-4} 
                  & \multicolumn{1}{c|}{\textbf{\textit{MIoU}}} & \multicolumn{1}{c|}{\textbf{\textit{Total ACC}}} &\textbf{\textit{AVG ACC}}   \\ \hline
                  Semantic NeRF& \multicolumn{1}{c|}{0.932} & \multicolumn{1}{c|}{0.991} &0.957  \\ \hline
                  Our Method& \multicolumn{1}{c|}{0.886} & \multicolumn{1}{c|}{0.983} & 0.930 \\ \hline
\end{tabular}
\label{tab1}
\end{center}
\end{table}

\subsubsection{Pixel-wise noise}
\par In some practical applications, some images may generate noise, making them unsuitable for direct use in training. The experiments in Semantic-NeRF \cite{zhi2021place} demonstrate a strong denoising capability, and we will demonstrate through similar experiments that our method also possesses this denoising ability. We set the experiment conditions so that Semantic-NeRF and our approach add the same noise to 50\% of the same pixels and 90\% of the same pixels, making the image blurry and then using the noisy images as a training set. 
\par From the results in Fig. 5, 50\% of noise label, but 90\% of noise label is hard to recognise by a human. Our method can still denoise the training set with noise and restore it to its original appearance. From the quantitative results in Table III, our results show its ability to denoise the semantic labels.

\begin{table}[!htbp]
\caption{QUANTITIVE RESULT IN PIXEL-WISE NOISE}
\begin{center}
\resizebox{\linewidth}{!}{
\begin{tabular}{|cl|c|ccc|}
\hline
\multicolumn{2}{|c|}{\multirow{2}{*}{\textbf{Model}}} & \multirow{2}{*}{\textbf{Nosie Ratio}} & \multicolumn{3}{c|}{\textbf{Metric}}                                                                               \\ \cline{4-6} 
\multicolumn{2}{|c|}{}                  &                   & \multicolumn{1}{c|}{\textbf{\textit{MIoU}}}                  & \multicolumn{1}{c|}{\textbf{\textit{Total ACC}}}                  & \textbf{\textit{AVG ACC}}                  \\ \hline
\multicolumn{2}{|c|}{\multirow{4}{*}{Semantic NeRF}} & \multirow{2}{*}{50\%} & \multicolumn{1}{c|}{\multirow{2}{*}{0.957}} & \multicolumn{1}{c|}{\multirow{2}{*}{0.994}} & \multirow{2}{*}{0.978} \\
\multicolumn{2}{|c|}{}                  &                  & \multicolumn{1}{c|}{}                  & \multicolumn{1}{c|}{}                  &                   \\ \cline{3-6} 
\multicolumn{2}{|c|}{}                  & \multirow{2}{*}{90\%} & \multicolumn{1}{c|}{\multirow{2}{*}{0.863}} & \multicolumn{1}{c|}{\multirow{2}{*}{0.990}} & \multirow{2}{*}{0.885} \\
\multicolumn{2}{|c|}{}                  &                   & \multicolumn{1}{c|}{}                  & \multicolumn{1}{c|}{}                  &                   \\ \hline
\multicolumn{2}{|c|}{\multirow{4}{*}{Our Method}} & \multirow{2}{*}{50\%} & \multicolumn{1}{c|}{\multirow{2}{*}{0.959}} & \multicolumn{1}{c|}{\multirow{2}{*}{0.994}} & \multirow{2}{*}{0.975} \\
\multicolumn{2}{|c|}{}                  &                   & \multicolumn{1}{c|}{}                  & \multicolumn{1}{c|}{}                  &                   \\ \cline{3-6} 
\multicolumn{2}{|c|}{}                  & \multirow{2}{*}{90\%} & \multicolumn{1}{c|}{\multirow{2}{*}{0.869}} & \multicolumn{1}{c|}{\multirow{2}{*}{0.987}} & \multirow{2}{*}{0.898} \\
\multicolumn{2}{|c|}{}                  &                   & \multicolumn{1}{c|}{}                  & \multicolumn{1}{c|}{}                  &                   \\ \hline
\end{tabular}
}
\end{center}
\end{table}

\begin{table}[!ht]
\caption{QUANTITIVE RESULT IN SUPER-RESOLUTION}
\begin{center}
\resizebox{\linewidth}{!}{
\begin{tabular}{|c|c|c|ccc|}
\hline
\multirow{2}{*}{\textbf{Model}}              & \multirow{2}{*}{\textbf{Down-Scale Method}}       & \multirow{2}{*}{\textbf{Down-Scale Factor}} & \multicolumn{3}{c|}{\textbf{Metric}}                            \\ \cline{4-6} 
                               &                         &                   & \multicolumn{1}{c|}{\textbf{\textit{MIoU}}} & \multicolumn{1}{c|}{\textbf{\textit{Total ACC}}} &\textbf{\textit{AVG ACC}}  \\ \hline
\multirow{4}{*}{Semantic-NeRF} & \multirow{2}{*}{Dense}  & 8                 & \multicolumn{1}{c|}{0.617} & \multicolumn{1}{c|}{0.895} &0.717  \\ \cline{3-6} 
                               &                         & 16                & \multicolumn{1}{c|}{0.431} & \multicolumn{1}{c|}{0.810} &0.537  \\ \cline{2-6} 
                               & \multirow{2}{*}{Sparse} & 8                 & \multicolumn{1}{c|}{0.901} & \multicolumn{1}{c|}{0.986} & 0.942 \\ \cline{3-6} 
                               &                         & 16                & \multicolumn{1}{c|}{0.826} & \multicolumn{1}{c|}{0.974} &0.886  \\ \hline
\multirow{4}{*}{Our Method}    & \multirow{2}{*}{Dense}  & 8                 & \multicolumn{1}{c|}{0.603} & \multicolumn{1}{c|}{0.890} &0.707  \\ \cline{3-6} 
                               &                         & 16                & \multicolumn{1}{c|}{0.423} & \multicolumn{1}{c|}{0.794} &0.535  \\ \cline{2-6} 
                               & \multirow{2}{*}{Sparse} & 8                 & \multicolumn{1}{c|}{0.881} & \multicolumn{1}{c|}{0.983} &0.925  \\ \cline{3-6} 
                               &                         & 16                & \multicolumn{1}{c|}{0.795} & \multicolumn{1}{c|}{0.968} &0.851  \\ \hline
\end{tabular}
}
\end{center}
\end{table}

\subsubsection{Super-resolution}
Super-resolution recovery plays a vital role in image processing, restoring low-pixel images to high-pixels through CNN methods such as \cite{zhang2018image}, \cite{wang2018esrgan}. Semantic-NeRF \cite{zhi2021place} demonstrates the ability of super-resolution recovery. The remaining part is used as training pixels after reducing the image by a particular proportion. The reduced part is used as a mask to allow the training process to restore the image to its original appearance. To verify that our method can achieve the above, we set the experiment as follows: 
\begin{itemize}
\item Shrink the image eight times, from 320x240 to 40x30, then scale back to the original size and only retain the semantic image information after reduction for training. 
\item Down-scaling the image 16 times, from 320x240 to 20x15, then scaling back to the original size and only retaining the semantic part information for training.
\end{itemize}

We use 'dense' and 'sparse' two down-scale methods for this process, ‘dense’ is we down-scale our input training image 8 or 16 times and then scale back by interpolating the nearest point value, ‘sparse’ is we make our input training image on every 8 or 16 girds corner retains the original values, and void all other points.
\par Fig. 6 shows the image after shrinking and restoring. Observing the image, our method can still perform super-resolution recovery 8 and 16 times down-scale in sparse label input. In the dense label input, in the 8 and 16 times down-scale, both methods also recover similar quality. This indicates that our method has a similar ability to Semantic-NeRF in the super-resolution method. The results in Table IV validate our claim.

\begin{table}[!ht]
\caption{QUANTITIVE RESULT IN LABEL PROPAGATION}
\begin{center}
\resizebox{\linewidth}{!}{
\begin{tabular}{|c|cl|ccc|}
\hline
\multirow{2}{*}{\textbf{Model}}              & \multicolumn{2}{c|}{\multirow{2}{*}{\textbf{Image Area \% per Class}}} & \multicolumn{3}{c|}{\textbf{Metirc}}                            \\ \cline{4-6} 
                               & \multicolumn{2}{c|}{}                  & \multicolumn{1}{c|}{\textbf{\textit{MIoU}}} & \multicolumn{1}{c|}{\textbf{\textit{Total ACC}}} &\textbf{\textit{AVG ACC}}  \\ \hline
\multirow{3}{*}{Semantic NeRF} & \multicolumn{2}{c|}{1\%}               & \multicolumn{1}{c|}{0.739} & \multicolumn{1}{c|}{0.946} &0.952  \\ \cline{2-6} 
                               & \multicolumn{2}{c|}{5\%}               & \multicolumn{1}{c|}{0.882} & \multicolumn{1}{c|}{0.977} &0.959  \\ \cline{2-6} 
                               & \multicolumn{2}{c|}{10\%}              & \multicolumn{1}{c|}{0.910} & \multicolumn{1}{c|}{0.984} & 0.959 \\ \hline
\multirow{3}{*}{Our Method}    & \multicolumn{2}{c|}{1\%}               & \multicolumn{1}{c|}{0.654} & \multicolumn{1}{c|}{0.900} &0.897  \\ \cline{2-6} 
                               & \multicolumn{2}{c|}{5\%}               & \multicolumn{1}{c|}{0.815} & \multicolumn{1}{c|}{0.948} &0.928  \\ \cline{2-6} 
                               & \multicolumn{2}{c|}{10\%}              & \multicolumn{1}{c|}{0.869} & \multicolumn{1}{c|}{0.967} &0.934  \\ \hline
\end{tabular}
}
\end{center}
\end{table}
\subsubsection{Label propagation}

Scene label annotation usually involves annotating specific labels in the image. These annotations provide essential metadata. Semantic-NeRF \cite{zhi2021place} concluded that this NeRF-based method could use a minimal number of small area blocks to recover the overall semantic label to ensure the accuracy and consistency of the data. The experimental setup is that in each semantic label corresponding to the image area, 1\%, 5\% and 10\% of the corresponding area are selected as our training set to enter our method and Semantic-NeRF training process to obtain results. Fig. 7 shows that our method and Semantic-NeRF can recover the original semantic label information within those selected areas. In Table V, the quantitative results show our method is not as good as the Semantic-NeRF.

\section{Conclusion}
In conclusion, this study demonstrates a novel approach using only semantic information. It has been observed that, even without reliance on RGB images, the proposed method effectively reconstructs scenes and generates semantic output, showcasing its robustness and adaptability. Our experiments illustrate that the method holds up well when handling noise, achieving super-resolution recovery and label propagation, demonstrating its resilience and applicability. An important observation was that our approach preserved multi-view consistency, which is crucial to 3D reconstruction methods. Furthermore, we demonstrated that this method could successfully train and deliver reliable results with only a fraction of the semantic keyframes. In future works, we will add more experiment 3D scenes to see the performance of our model and find some method to reduce computational loss.




\begin{figure}[ht]
\centerline{\includegraphics[width=0.5\textwidth]{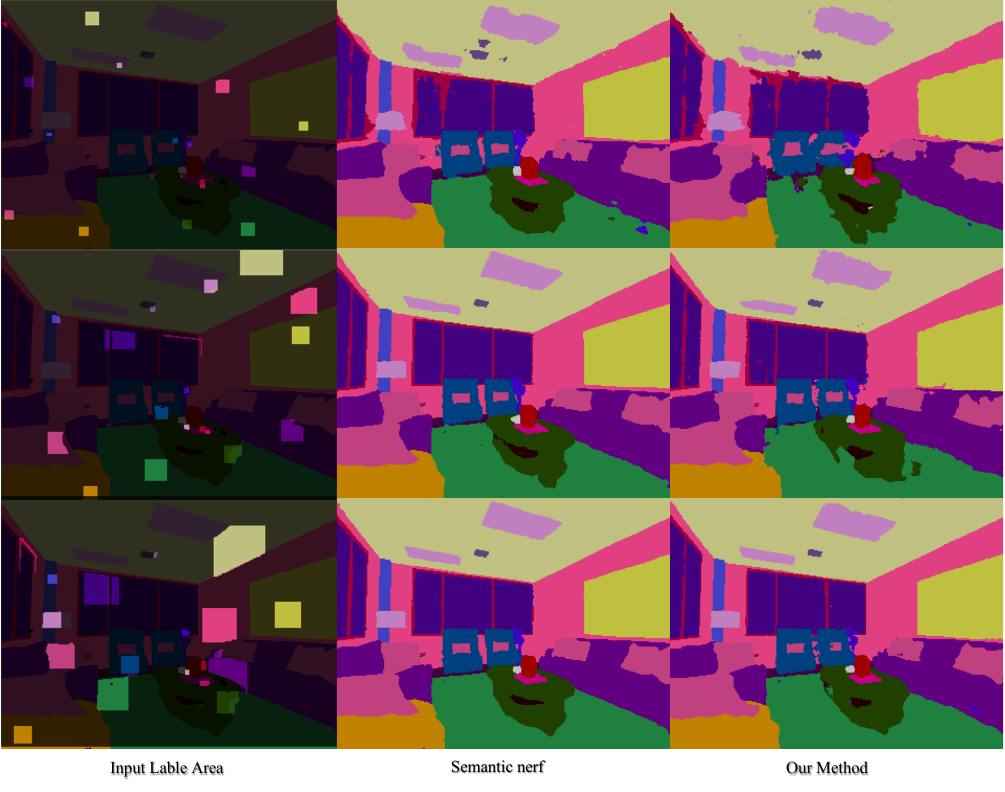}}
\caption{Qualitative comparison of label propagation. In the training step, we use each image's partial semantic input area with 1\%, 5\% and 10\%. In each row representing 1\% area, 5\% area and 10\% area respectively, our method can render the semantic output with partial label compared to Semantic-NeRF.}
\label{fig}
\end{figure}

\bibliographystyle{IEEEtran}
\bibliography{IEEEabrv,nerf_manual}

\end{document}